\documentclass[twoside,runningheads]{llncs}

\usepackage{citefoot}

\usepackage{graphicx}

\usepackage{amsmath}

\usepackage{paralist} 

\usepackage{hyperref}
\usepackage{xcolor}

\usepackage{xskak,chessboard}

\usepackage{msymb,lsyntax}

\usepackage{prettyref}
\newcommand{\rref}[2][]{\prettyref{#2}}
\newrefformat{sec}{Section\,\ref{#1}}
\newrefformat{def}{Def.\,\ref{#1}}
\newrefformat{thm}{Theorem\,\ref{#1}}
\newrefformat{prop}{Proposition\,\ref{#1}}
\newrefformat{lem}{Lemma\,\ref{#1}}
\newrefformat{cor}{Corollary\,\ref{#1}}
\newrefformat{ex}{Example\,\ref{#1}}
\newrefformat{tab}{Table\,\ref{#1}}
\newrefformat{fig}{Fig.\,\ref{#1}}
\newrefformat{app}{Appendix\,\ref{#1}}

\usepackage{ifpdf}
\ifpdf
\fi

\begin{document}

\title{Intersymbolic AI\thanks{
Funding has been provided by an Alexander von Humboldt Professorship for AI and the Helmholtz KiKIT Kerninformatik / Core Informatics at KIT.}}
\subtitle{Interlinking Symbolic AI and Subsymbolic AI}
\author{Andr\'e Platzer\orcidID{0000-0001-7238-5710}}
\institute{
  Karlsruhe Institute of Technology, Karlsruhe, Germany
  \email{platzer@kit.edu}
}
\titlerunning{Intersymbolic AI: Interlinking Symbolic AI and Subsymbolic AI}

\maketitle

\begin{abstract}
This perspective piece calls for the study of the new field of \emph{Intersymbolic AI}, by which we mean the combination of \emph{symbolic AI}, whose building blocks have inherent significance/meaning, with \emph{subsymbolic AI}, whose entirety creates significance/effect despite the fact that individual building blocks escape meaning.
Canonical kinds of symbolic AI are logic, games and planning.
Canonical kinds of subsymbolic AI are (un)supervised machine and reinforcement learning.
Intersymbolic AI interlinks the worlds of symbolic AI with its compositional symbolic significance and meaning and of subsymbolic AI with its summative significance or effect to enable culminations of insights from both worlds by going between and across symbolic AI insights with subsymbolic AI techniques that are being helped by symbolic AI principles.
For example, Intersymbolic AI may start with symbolic AI to understand a dynamic system, continue with subsymbolic AI to learn its control, and end with symbolic AI to safely use the outcome of the learned subsymbolic AI controller in the dynamic system.
The way Intersymbolic AI combines both symbolic and subsymbolic AI to increase the effectiveness of AI compared to either kind of AI alone is likened to the way that the combination of both conscious and subconscious thought increases the effectiveness of human thought compared to either kind of thought alone.
Some successful contributions to the Intersymbolic AI paradigm are surveyed here but many more are considered possible by advancing Intersymbolic AI.

\end{abstract}

\section{Introduction}

\emph{Artificial Intelligence (AI)} has received both waves of significant attention and of significant success \cite{RusselNorvig:21}.
AI comes in two flavors: symbolic and subsymbolic AI.
\emph{Symbolic AI} \cite{DBLP:journals/cacm/NewellS76} is ultimately rooted in symbolic techniques whose individual building blocks carry meaning and are, thus, interpretable (at least in principle although not necessarily at the full required scale) and where individual building blocks are typically (de)composed to obtain a whole.
\emph{Subsymbolic AI} is ultimately rooted in numerical techniques where the individual building blocks do not carry direct meaning (beyond the role they happen to play in the context of the computation) and where the result is given indirectly, e.g., by an iterative optimization procedure fed from training data or experience.
Both flavors of AI have remarkably different and sometimes quite complementary strengths and weaknesses.
Their combination in what we call \emph{Intersymbolic AI}, thus, has the potential to add the strengths of symbolic AI and of subsymbolic AI while canceling out the weaknesses of symbolic AI and of subsymbolic AI, respectively.

The difference is in significance.
\emph{Symbolic AI} studies AI whose building blocks and operating principles carry meaning, which makes them compositionally significant in a semiotic sense \cite{Eco76}.
\emph{Subsymbolic AI} studies AI whose building blocks do not carry individual clear meaning and yet whose overall outcome carries (phenomenologically) pragmatic significance or effect.
\emph{Intersymbolic AI} interlinks symbolic AI and subsymbolic AI such that some building blocks and operating principles carry meaning, while others do not, in ways to achieve overall outcomes of increased significance.
Just hoping that the respective advantages of symbolic and subsymbolic AI are additive while their respective weaknesses naturally cancel each other out is, of course, na\"ive.
But careful combinations of symbolic and subsymbolic AI fulfill this promise of the best of both worlds.

One analogy why Intersymbolic AI combinations of symbolic AI and subsymbolic AI can be more effective than either part alone is similar to how combinations of conscious and subconscious thought can increase the effectiveness of human thought.
Just like symbolic AI, conscious thought seems more symbolic and can be explained better compared to subconscious thought that seems harder to give direct meaning to or explain, even if this is elusive to know \cite{Popper62}.
Humans combine both thought processes effectively, which is why such combinations were proposed as analogies for cognitive architectures \cite{DBLP:conf/bica/SukhobokovGC19}.

Canonical examples of symbolic AI include logic \cite{DBLP:journals/tit/NewellS56,DBLP:conf/ifip/NewellSS59}, planning \cite{DBLP:series/synthesis/2013Geffner}, and game play \cite{DBLP:journals/ibmrd/NewellSS58,Hsu02}.
Canonical examples of subsymbolic AI include neural networks \cite{RusselNorvig:21} and reinforcement learning \cite{SuttonBarto18}.
Probabilistic reasoning is a canonical example but either falls on the symbolic AI side of Bayesian inference in Bayesian networks \cite{DBLP:books/daglib/0066829,RusselNorvig:21} or falls on the subsymbolic AI side of Bayesian statistics and Markov chain Monte Carlo \cite{GaelFR24}.
Due to the fundamentally different characteristics, advantages, and disadvantages of many forms of symbolic AI compared to many forms of subsymbolic AI, the combination of symbolic AI and subsymbolic AI in intersymbolic AI has much to gain if exercised with suitable care.
The weights at individual neurons in a neural network are hard to interpret, while each operator in a logical formula is easily interpreted.
Conversely, neural networks have a simple and powerful mechanism for learning how they best reproduce training data \cite{RusselNorvig:21} while logical formulas are harder to learn \cite{DBLP:journals/iandc/Angluin87} and it is, sadly, challenging to integrate logical knowledge into neural networks \cite{TowellS94}.

Successful combinations contributing to Intersymbolic AI include a range:%
\begin{enumerate}
\item \label{item:SafeAI4CPS}
combining reinforcement learning with hybrid systems theorem proving and proof-based synthesis to enable Safe AI in CPS \cite{DBLP:conf/aaai/FultonP18,DBLP:conf/tacas/FultonP19,DBLP:conf/itc/FultonP18,DBLP:conf/qest/Platzer19},

\item \label{item:Pegasus}
combining symbolic proof, first integrals and Darboux polynomials with eigensystems, sum-of-squares programming and linear programming for invariant generation of differential equations \cite{DBLP:journals/fmsd/SogokonMTCP22},

\item \label{item:Looprl}
combining AlphaZero for reinforcement learning on deep neural networks with theorem proving and nondeterministic programming for loop invariant synthesis \cite{DBLP:conf/nips/LaurentP22},

\item \label{item:waypoints}
combining neural network path tracking and model-predictive control with hybrid systems theorem proving for safe waypoint following vehicles \cite{DBLP:journals/csyl/LinMPD22},

\item \label{item:CESAR}
combining arithmetic simplification heuristics and refinement approximations with hybrid systems verification and game theory for control envelope synthesis \cite{DBLP:conf/tacas/KabraLMP24},

\item \label{item:VerSAILLE}
combining neural network verification with complete arithmetic linearization techniques and hybrid systems theorem proving for safety verification of Neural Network Control Systems \cite{DBLP:journals/corr/abs-2402-10998},

\item combining Graph Neural Networks with symbolic proof to help automate higher-order logic provers \cite{DBLP:conf/aaai/PaliwalLRBS20,DBLP:conf/icml/BansalLRSW19}.
\end{enumerate}
While quite different, these uses all share the design principle that symbolic AI principles are combined with subsymbolic techniques.
While the relative proportion of the use of symbolic AI versus subsymbolic AI in these combinations is different, this range shows the potential of Intersymbolic AI for vastly different purposes.
Two technical connections arise in many of these Intersymbolic AI uses, dynamic logic proving ideas \cite{DBLP:conf/lics/Platzer12a} arise in combinations \ref{item:SafeAI4CPS}--\ref{item:VerSAILLE} and ideas from the shielding technique ModelPlex \cite{DBLP:journals/fmsd/MitschP16} arise in combinations \ref{item:SafeAI4CPS} and \ref{item:waypoints}--\ref{item:VerSAILLE}.
But many other combinations contributing to Intersymbolic AI are conceivable more broadly as well, which is why this perspective piece is meant as a call to the scientific community to notice the challenge and contribute to its solution.
The record here is slanted by the author's prior specific experience with manifestations of Intersymbolic AI, but the general phenomenon of Intersymbolic  AI is by no means meant to be understood exclusive to the existing combinations.

This paper is not the first calling for distinctions and combinations of different kinds of AI (good old fashioned AI \cite{Haugeland89}, neuro-symbolic AI and related notions \cite{DBLP:series/cogtech/GarcezLG2009,DBLP:conf/aaai/BoochFHKLLLMMRS21,DBLP:journals/aim/Kautz22}, considering such combinations the most important AI challenges \cite{DBLP:journals/bams/Avigad24}) and it certainly will not be the last.\footnote{%
The difference between symbolic AI versus subsymbolic AI has also sometimes been sought in the contrast of traditional versus nontraditional AI, or in rule-based versus learning-based AI, or in good old fashioned AI \cite{Haugeland89} versus neural networks.
Symbolic AI versus subsymbolic AI better characterizes the essential source of the difference.
}
The author's hope is to contribute to the state of understanding by better identifying the logical root cause for the divide between symbolic versus subsymbolic AI, calling for a pervasive and significantly more broadly interpreted combination in the form of Intersymbolic AI, and inspiring by showcasing a wide range of successful novel styles of achieving Intersymbolic AI with entirely different flavors of combinations.

The plan for this perspective piece is to characterize typical symbolic AI in \rref{sec:symbolicAI}, subsymbolic AI in \rref{sec:subsymbolicAI}, then contrast and compare their strengths leading to the call for intersymbolic AI in \rref{sec:intersymbolicAI}.
This paper will then make a short foray into one particularly compelling application area for mixing intersymbolic AI in cyber-physical systems in \rref{sec:IntersymbolicAIinCPS}, and give a short overview of successful intersymbolic AI uses in \rref{sec:intersymbolicAI-survey} to demonstrate its wide range before concluding in \rref{sec:Conclusion}.

\section{Symbolic AI} \label{sec:symbolicAI}

Symbolic AI was pioneered already in its purest form of AI by deductive proof in the first AI tools, the Logic Theory Machine and General Problem Solver, in the 1950s by Allen Newell and Herbert Simon \cite{DBLP:journals/tit/NewellS56,DBLP:conf/ifip/NewellSS59} based on traditional logic \cite{Frege79,Goedel30,Gentzen35I}.
Even the origin of logic such as in Aristotle's syllogisms, Frege's \emph{Begriffsschrift} \cite{Frege79}, and Gentzen's natural deduction \cite{Gentzen35I,DBLP:journals/jsyml/Quine50} were already meant to rigorously capture the human process of argumentative thought and reasoning.
Since every symbol and operator in logic carries intrinsic meaning, and since deductive proof preserves meaning, this pure style of symbolic AI has the clearest significance (in the semiotic sense \cite{Eco76,Sowa00} that its parts and outcomes are meaningful), the clearest interpretations (originating from the compositions of the meanings of the individual symbols and deductive proof rules), and the clearest explanations (the justification in the form of the resulting deductive proof).

Logical inferences such as the reasoning captured in the logical formula
\begin{equation}
(\textrm{rain}\limply\textrm{wet}) \land \lnot\textrm{wet} \limply \lnot\textrm{rain}
\label{eq:fmlexample}
\end{equation}
can be justified by deductive proof and carry clear meaning independent from the question what exactly one means with raining and what exactly one means by a wet road.
The inference stands and can be justified just from the logical structure of the formula \rref{eq:fmlexample} and the meaning of its logical connectives $\land,\lnot,\limply$.
In an application, formula \rref{eq:fmlexample} can be read to say that if an AI assumes that the road is wet when it is raining, but the AI observes that the road is not wet, then the AI concludes that it cannot be raining.
This is an undeniably correct inference.
Of course, the observation from a visual image that the road is, indeed, not wet is an entirely different matter, and better handled using subsymbolic AI in the form of neural networks for image classification.
The main mode of using logic is by deductive proof \cite{Fitting96a,DBLP:journals/logcom/BachmairG94,DBLP:books/el/RobinsonV01,Platzer10} or satisfiability modulo theory solving \cite{DBLP:journals/jacm/NieuwenhuisOT06}, but some uses include symbolic learning \cite{DBLP:journals/iandc/Angluin87,DBLP:conf/ijcai/CropperDM20}.

Symbolic AI by planning \cite{DBLP:conf/ijcai/FikesN71,DBLP:series/synthesis/2013Geffner} and game play \cite{DBLP:journals/ibmrd/NewellSS58} are also such that the individual actions in a plan or game have clear symbolic and meaningful character, where it is clear what these building blocks mean and what the effect of taking those actions is.
Even in games where the actual outcome of an action partially depends on chance, the game itself is still described in a meaningful way, which is ultimately the prerequisite to solving it strategically \cite{vonNeumannMorgenstern}, because it is hard to play a game strategically that one does not understand.
Likewise, planning requires a clear understanding of the meaning and effect of actions, otherwise a plan would reduce to mere guesswork, which is the exact opposite of a rational plan.
This principle continues to apply in the significantly more challenging field of planning under uncertainty \cite{DBLP:series/synthesis/2013Geffner}, which, despite the uncertainty, still necessitates a clear understanding of the meaning of each action, as well as of the remaining uncertainty about its outcome and/or the interference by the environment.
Several different approaches have been introduced for representing knowledge about actions for planning, including situation calculus \cite{DBLP:conf/birthday/Reiter91,DBLP:conf/kr/Reiter96}, fluent calculus \cite{DBLP:journals/tplp/Thielscher05}, event calculus \cite{DBLP:books/daglib/0095085}, see \cite{DBLP:journals/csur/MitschPRS15} for a survey.

\begin{figure}[tbhp]
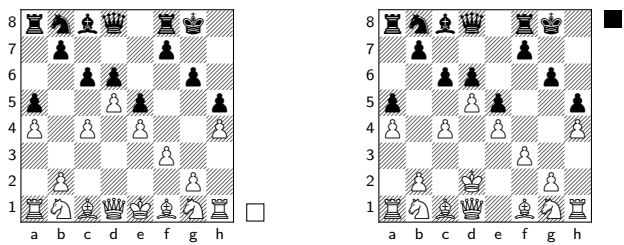

\centering
\begin{minipage}{4cm}
\setchessboard{smallboard,showmover=true,boardfontsize=10pt,labelfontsize=6pt}%
\newchessgame[white=Paethz,black=Dirr,result=0-1]%
\hidemoves{%
1.d4 Sf6 2.c4 g6 3.Sc3 Lg7 4.e4 d6 5.Sge2 O-O 6.Sg3 c6 7.Le2 a6 8.a4 a5 9.h4 h5 10.Le3 Sa6 11.f3 e5 12.d5 Sd7 13.Sf1 Sdc5 14.Sd2 Db6 15.Db1 Sb4 16.Sb3 Scd3+}
    \chessboard%
\end{minipage}
\qquad%
\begin{minipage}{4cm}
\setchessboard{smallboard,showmover=true,boardfontsize=10pt,labelfontsize=6pt}%
\newchessgame[white=Paethz,black=Dirr,result=0-1]%
\hidemoves{%
1.d4 Sf6 2.c4 g6 3.Sc3 Lg7 4.e4 d6 5.Sge2 O-O 6.Sg3 c6 7.Le2 a6 8.a4 a5 9.h4 h5 10.Le3 Sa6 11.f3 e5 12.d5 Sd7 13.Sf1 Sdc5 14.Sd2 Db6 15.Db1 Sb4 16.Sb3 Scd3+ 17.Kd2}
    \chessboard%
\end{minipage}
  \caption{A move of the white king at \texttt{e1} to \texttt{d2} in a chess play}
  \label{fig:exchess}
\end{figure}

It takes sufficient symbolic understanding of the rules of chess to find that the move of the white king from field \texttt{e1} to \texttt{d2} in \rref{fig:exchess}, for example, is a legal move.
Without an understanding of the game at least in the form of which moves are legal when, and what the objective of the game is in the first place, no strategic game play beyond random trial and error is possible.
Symbolic game knowledge includes the fact that kings can move one field to any of the up to 8 adjacent fields not occupied by pieces of the same player and that the objective is to capture the opponent's king without losing your own.
After a computer program won checkers, which much later led to solving the game of checkers with perfect game plays \cite{SchaefferBBKMLLS07}, Deep Blue was the first computer to defeat the human world chess champion \cite{Hsu02}.
It worked by incredibly efficient heuristic search \cite{DBLP:books/daglib/0068933} in games enabling the exploration possibilities many moves into the future despite the resulting combinatorial explosion.

Despite an initial excitement about logic and game search based symbolic AI and additional waves of symbolic AI such as in so-called expert systems \cite{DBLP:journals/cacm/Feigenbaum96} and AI planning \cite{GhallabNT16}, purely symbolic AI also has its shortcomings that were initially underestimated considerably.
It is, retroactively somewhat unsurprisingly, exceedingly difficult to exhaustively formalize all required facts about the world to enable general common-sense reasoning.
And, yet, symbolic AI has achieved impressive feats that go beyond solving chess \cite{DBLP:books/daglib/0068933}, and use logic to prove new theorems in Boolean algebra \cite{DBLP:journals/jar/McCune97}, verify hardware and software \cite{DBLP:journals/cacm/ClarkeES09}, solve extraordinarily large SAT instances \cite{DBLP:series/faia/336}, find bugs and prove fixes in the Java collections library \cite{DBLP:conf/cav/GouwRBBH15}, or prove safety and find bugs in the FAA's Next-generation Airborne Collision Avoidance System ACAS~X \cite{DBLP:journals/sttt/JeanninGKSGMP17}, formalize the proof of the Four Color Theorem \cite{Gonthier05}, the prime number theorem \cite{DBLP:journals/tocl/AvigadDGR07}, and the Feit-Thompson theorem on the solvability of finite groups of odd order \cite{DBLP:conf/itp/GonthierAABCGRMOBPRSTT13}, or prove the 400 year old Kepler conjecture on sphere packing \cite{Hales17}.

\section{Subsymbolic AI} \label{sec:subsymbolicAI}

The most canonical example of subsymbolic AI is that of a neural network (NN) \cite{RusselNorvig:21}, including Convolutional Neural Networks (CNNs) \cite{LeCunBD89}, for machine learning \cite{Mitchell:240}.
In its basic form, a (feed-forward) NN is a way of representing a real vectorial function via a fixed finite sequence of layers
\begin{equation}
x^{(n+1)}=f^{(n)}\big(A^{(n)}x^{(n)}+b^{(n)}\big)
\label{eq:forward-NN}
\end{equation}
of real vectors $x^{(n)}$ with fixed nonlinear activation functions $f^{(n)}:\reals\to\reals$ and matrices $A^{(n)}$ of real-valued weights and vectors $b^{(n)}$ of real-valued offsets (of compatible dimensions) describing the linear transformation \(A^{(n)}x^{(n)}+b^{(n)}\).
Learning or training an NN consists of backpropagation via the partial derivatives of \rref{eq:forward-NN} to optimize $A^{(n)}$ and $b^{(n)}$ so that they better fit to a given training set of expected input/output data (starting from the final output at the last $n$).
Even if the data propagation \rref{eq:forward-NN} as well as its backward propagation for training have easily comprehensible computations, the individual weights as well as the entirety of a NN is nearly impossible to give a clear meaning to, especially given the size of deep neural networks as in Large Language Models \cite{DBLP:conf/nips/VaswaniSPUJGKP17} which presently already have up to a trillion weights.
NNs are particularly successful for supervised learning where an expected input/output function is being learned based on a significant amount of training data.

If a sufficiently large number of given training images are labeled raining or not raining, then NN training leads to NNs that can classify other images effectively provided the original training data was representative for the real situations.
Of course, if all training images with rain were taken at night while all training images without rain were taken during daytime, then the NN might classify all training data correctly even if it accidentally learned the concept of nighttime instead of rain and, thus, generalizes poorly.
This is the difficult phenomenon of a distribution shift that fools learning when the distribution of examples during learning is not representative of the distribution experienced in reality.
Splitting off a small validation data set from the training data, which is usually done to avoid overfitting that merely memorizes the available data without generalization capabilities, does not help alleviate those concerns, because the validation data is chosen from the same distribution as the training data.

Reinforcement learning (RL) \cite{SuttonBarto18} is another canonical example of subsymbolic AI bridging ideas of machine learning with game theory to learn how to act in a game scenario from experience.
The basic idea is to repeatedly act according to some fixed policy, observe the outcome, and increase the probability with which the policy takes that action in the previous state if the actual outcome is favorable, otherwise decrease it.
For actions that directly reach a terminal state in which the assigned outcome can be read off, this is easy to tell.
For actions that do not reach a terminal state, their future outcome is uncertain so that an estimate of the expected payoff based on the present experience reflected in the present policy is used instead.
In its basic form, it suffices to give RL agents an experiential black box access to the system they are trying to learn how to control, without the need for a symbolic understanding of its pieces and their meanings.
RL is very flexible for a range of different problems but is also not particularly sample-efficient.
That is, it takes a very large number of experiments for RL to actually learn.

To understand why RL is sample-inefficient, imagine having to learn chess strategies by repeatedly taking random actions on a chess board and observing whether you've won or lost in the end.
It is not at all easy to trace this ultimate fate back to the impact of the individual move decisions along the way (such as the one in \rref{fig:exchess}) and takes an absurdly large number of practice plays to learn meaningful generalizable strategies this way.
And yet, this is what the AlphaZero algorithm ultimately achieved for games including chess and Go using a massive number of practice rounds by combining Monte Carlo Tree Search (MCTS) and subsymbolic AI of DNNs \cite{DBLP:journals/nature/SilverSSAHGHBLB17,DBLP:journals/corr/abs-1712-01815}.
AlphaZero is already combining symbolic and subsymbolic AI principles even if its pieces are still further away from the compositional and interpretable aspects of canonical symbolic AI.

The blackbox access of RL to the system is simultaneously its blessing and its curse.
The advantage is precisely that not much is needed to use RL.
The downside is that not much is known about the resulting policy.
For principle reasons, at most statistical guarantees can be provided for every blackbox numerical access even to fairly tame continuous systems \cite{DBLP:conf/hybrid/PlatzerC07}.
For finite state systems with finite actions, guarantees can still be given in the limit if the (initial) policy never considers any actions entirely impossible \cite{SuttonBarto18}.
The reason for the latter requirement is exactly that the only good action could otherwise have been considered impossible initially and would, thus, almost surely never even have been tried.

The performance demonstrated by subsymbolic AI techniques is impressive, but they are subsymbolic since one would be hard-pressed to give the meaning and justification for one individual weight picked arbitrarily in the middle of an underlying NN like \rref{eq:forward-NN}, for example.
This, after all, is not their working principle.
Rather, the overall outputs of an NN and the decisions of an RL agent are a direct function of the examples and experiences they have seen during training.
More to the point, it is exceedingly difficult to know whether the particular answer generated by subsymbolic AI is the right one \cite{DBLP:journals/corr/AmodeiOSCSM16,DBLP:conf/fedcsis/KwiatkowskaZ23}.
The working principle of subsymbolic AI, after all, is not one of a composition of lots of little parts with clear meaning as is the case in symbolic AI such as logic.

\section{Intersymbolic AI} \label{sec:intersymbolicAI}

Both symbolic AI and subsymbolic AI are useful and have been demonstrated to solve impressive challenges, even if both also have limitations.
In a nutshell, exemplified for emphasis in the case of its purest form of logic, \emph{symbolic AI}:
\begin{enumerate}
\item has high accuracy and precision:
logical inferences follow sound proof rules that only ever draw correct conclusions from correct premises.
While a proof search does not have to succeed reliably if the proof search procedure is not deterministic, the proof always justifies correct consequences of the premises.

\item is self-explanatory:
logical symbols carry clear meaning that, by the logical principle of compositionality, give meaning to any combination of them.
Moreover, a proof of a logical formula gives syntactic meaning to and explains the formula in a way that can be syntactically introspected (which is particularly apparent in constructive logic \cite{Godel32,MartinLof80}).

\item depends on manual effort to formalize:
one of the biggest strengths of logic is also its biggest weakness, the fact that logic is ontologically neutral.
The rules of logic do not depend on whatever the specific application is.
Having said that, application-specific conclusions then depend on formalizing the underlying principles as formulas in the logic that can be used as premises to draw interesting conclusions from.
Knowledge representation is the area of symbolic AI devoted to this nontrivial challenge \cite{Sowa00,BrachmanLevesque14,DBLP:journals/csur/MitschPRS15}.

\item is better at correct reasoning than at common-sense reasoning:
it is precisely logic's insistence on correctness that holds it back in common-sense reasoning.
In fact, symbolic AI would be perfect at common-sense reasoning, but only after the nearly infeasible challenge of formalizing all relevant parts of the world to obtain the required common-sense knowledge.
And even if that were possible, fast symbolic reasoning is difficult to achieve in huge knowledge bases without useful structures \cite{DBLP:books/daglib/0041477}.
Besides, a typical common-sense conclusion of the fact that Tweety is a bird is that Tweety flies. Except that this is, of course, an incorrect conclusion even if Tweety is a bird, in case Tweety is a penguin.
Except that this is also wrong in case Tweety the penguin found a jet pack.
Nonomonotonic logics partially alleviate this curse but also lack some of the strengths of classical logics \cite{Geffner:thesis,Lukaszewicz90,Brewka91,DBLP:journals/jacm/EiterG95}.

\item is challenging at scale:
while incredibly big and complicated proofs have been conducted in logic and logic-based reasoning \cite{Gonthier05,DBLP:journals/cacm/ClarkeES09,DBLP:series/faia/336,DBLP:conf/itp/GonthierAABCGRMOBPRSTT13,DBLP:journals/sttt/JeanninGKSGMP17,Hales17}, proof search is, nevertheless, challenging for complicated questions.
The fact that logic is symbolic so that all of its parts carry meaning and that logic naturally supports explanations makes sure that humans can help in interactive proof \cite{Gonthier05,DBLP:conf/itp/GonthierAABCGRMOBPRSTT13,DBLP:journals/sttt/JeanninGKSGMP17,Hales17}.
But this is both a blessing and a curse, because, for sufficiently challenging problems, the ability of humans to help also comes with the need for humans to help.
Proving the Kepler conjecture without human help is not realistic. But a proof of the Kepler conjecture was possible with the combined help of automatic symbolic logic and manual human effort \cite{Hales17}.
\end{enumerate}

\noindent
\emph{Subsymbolic AI} has mostly complementary strengths and weaknesses.
In a nutshell, exemplified for emphasis in the case of neural networks, \emph{subsymbolic AI}:
\begin{enumerate}
\item works without explicit manual programming:
even if a lot of art, cleverness, and manual effort is in the design of a neural network architecture its input features and in the suitable selection of training data as well as the hyper parameter tuning, the actual training runs are fully automatic and conveniently implemented on highly efficient libraries with GPU bindings such as TensorFlow \cite{DBLP:journals/corr/AbadiABBCCCDDDG16} and PyTorch \cite{DBLP:conf/nips/PaszkeGMLBCKLGA19}.

\item scales computationally:
the individual operations of a learning algorithm parallelize well and the outcome (approximately) fits the given data better the more computation time is spent learning from the available data according to experimental learning rate curves that are approximately linear in the data \cite{DBLP:journals/corr/abs-2001-08361}.
This is in contrast to symbolic AI, where spending more compute time does not make a proof search succeed any better if it used the entirely wrong proof strategy to begin with.
Parallel computation still helps find proofs if different proof strategies are used in clever ways or proofs are parallelized across separate subgoals \cite{DBLP:conf/itp/Wenzel13,DBLP:conf/icfem/RenshawLP11,DBLP:conf/cade/RegerTV15,DBLP:books/sp/HS2018,DBLP:conf/sat/SchreiberS21}.
But, except for very few special fragments, logical proof search techniques are at least exponential or NP-hard or even undecidable \cite{DBLP:conf/stoc/Cook71,DBLP:journals/bsl/Urquhart95,DBLP:journals/jacm/AtseriasM20,DBLP:conf/stoc/RezendeGNPR021}.

\item depends on absurd amounts of data:
unlike humans with their (sometimes fallible) generalization bias \cite{DBLP:journals/cogsci/PetersKB22}, machine learning techniques need absurd amounts of training data \cite{DBLP:conf/nips/VaswaniSPUJGKP17}, but then tend to tolerate errors and generalize well if the training distribution meets the real-life distribution.

\item is difficult to get correct:
the other side of the coin that subsymbolic AI does not \emph{need} explicit programming since its behavior is defined implicitly from data is that it does not \emph{support} explicit programming so its behavior is merely implicit and not explicitly clear either.
That is why there are numerous challenges to making AI safe \cite{DBLP:journals/corr/AmodeiOSCSM16,DBLP:conf/itc/FultonP18,DBLP:journals/arcras/KwiatkowskaN022,DBLP:conf/fedcsis/KwiatkowskaZ23}.
The unpredictability is particularly challenging in generative AI, such as represented by Generative Adversarial Networks (GAN) \cite{DBLP:journals/cacm/GoodfellowPMXWO20} and Large Language Models (LLMs) \cite{DBLP:conf/fat/GanguliHLABCCDD22}.

\item is hard to explain:
as made prominent with the area of eXplainable AI (XAI), subsymbolic AI techniques are famously at odds with explanations and even XAI's explanation mechanisms in image classification are easily fooled \cite{DBLP:conf/sp/NoppelPW23}.

\end{enumerate}

This complementary distribution of strengths and weaknesses calls for a synergy of symbolic AI with subsymbolic AI called Intersymbolic AI.

\emph{Intersymbolic AI} interlinks symbolic AI and subsymbolic AI such that some of its building blocks carry semantic significance while others remain subsymbolic creating effectful significance only in the composition, not in the parts.
While not all parts of intersymbolic AI are meaningful, its symbolic parts are and can carry and preserve a clear semantics within those parts, rendering itself to natural explanations.
While not all parts of intersymbolic AI are held to the principles of careful compositional semantical design, its subsymbolic parts can benefit from indirect characterizations of optimization such as end-to-end expectations in machine learning.
While not all parts of intersymbolic AI have equally justifiable outcomes, its symbolic parts provide clear correctness arguments.
While not all parts of intersymbolic AI need explicit characterizations of common-sense reasoning aspects, its subsymbolic parts can generalize (within distribution) from the knowledge that is so hard to capture explicitly except in experience.
While not all parts of intersymbolic AI are explainable, its symbolic parts provide natural mechanisms for explainability stemming from the symbolic semantics.

Intersymbolic AI supports the mode of operation where some of its parts obey the symbolic principles of compositional semantic meaning and obtain the resulting trust, while other parts remain opaque and sidestep the challenging questions of interpretability and trust by remaining subsymbolic.
Intersymbolic AI does not have to choose one of the above two extremes but can go in between, enabling combinations of techniques with complementary strengths.
By clever strategic choices which parts of intersymbolic AI best operate symbolically and which parts best operate subsymbolically, the respective intrinsic advantages of both arise naturally in the respective intersymbolic AI parts.
Canonical intersymbolic AI features symbolic AI in parts where a clear compositional meaning can be obtained or in which correct outcomes are crucial, and features subsymbolic AI in parts where crystal clear definitions of expected behavior are challenging or in which explicit solutions are difficult or impractical to obtain and implicit optimization outcomes are better.
Other combinations of symbolic and subsymbolic AI are perfectly conceivable for intersymbolic AI, however.

\section{A Case for Intersymbolic AI in Cyber-Physical Systems} \label{sec:IntersymbolicAIinCPS}

One area with a particularly obvious need for mixing symbolic AI and subsymbolic AI to form Intersymbolic AI is that of Autonomous Cyber-Physical Systems.
\emph{Cyber-Physical Systems (CPS)} combine cyber capabilities such as computation with physical capabilities such as the motion of a car or robot or the flight of an aircraft \cite{Platzer10,LeeSeshia13,Alur15,Platzer18,Mitra21,Marwedel21}.
The fact that CPS are safety-critical is the reason behind the well-deserved attention of symbolic AI generalizations to cover CPS correctness \cite{Platzer10,Alur15,Platzer18,Mitra21}.

In a pursuit to enable more autonomy, i.e., the self-propelled operation of CPS without external influence or extensive supervision by humans, \emph{Autonomous CPSs} see an increased use of subsymbolic AI, such as machine learning with neural networks, because that increases the flexibility and lessens the need for explicit manual programming.
But the challenge is that this use of subsymbolic AI in Autonomous CPS significantly complicates safety, exactly because the CPS is no longer programmed explicitly but is partially optimized or learned from data.
Of course, autonomy can also be achieved through careful design in sufficiently narrow domains, but the use of learning techniques precisely has the appeal that not all situations need to be considered explicitly, because of the generalization capabilities of learning techniques.
Despite a lot of progress in specific domains, it has proven exceedingly difficult to write programs for image recognition without the use of learning from training data, for instance.
As the development of the CPS moves from being based on intent (in the form of a manually written program) to experience (in the form of training data), it gets increasingly difficult, however, to then justify why the resulting autonomous CPS is safe to use.

More generally, while AI-based autonomy is, in some contexts, the only plausible route toward achieving system operability, the resulting lack of predictability is more detrimental to required safety guarantees, e.g., in self-driving cars, autonomous delivery drones, or mobile robots.
Limits in communication, situational awareness, and scale require autonomous operation for a sufficient duration until a human driver, pilot, or operator can step in safely.
AI helps adapt to situations flexibly without requiring explicit programming for all situations.
The ironic Catch-22 is that purely data-driven subsymbolic AI technologies fundamentally lack the guarantees and predictions required for the safe operation of a CPS. 
This is where a solid logical foundation for Safe AI in CPS \cite{DBLP:conf/itc/FultonP18,DBLP:conf/qest/Platzer19,DBLP:journals/corr/abs-2402-10998} develops intersymbolic AI combining symbolic and subsymbolic AI aspects to get the best of both worlds: compelling predictions that symbolic AI proof techniques provide alongside compelling flexibility that subsymbolic AI provides.

While it is fairly obvious that a supervisor could interfere with an AI or learning agent whenever those go astray, its crucial requirement for CPSs is that a safety analysis has to identify what exact interference is needed and when in order to guarantee a safe operation of the AI in the CPS going forward.
It is also important that all interference minimizes its impact on the AI's learning capabilities.
Otherwise, for instance, an excessive safety net might lead to a trivial learning outcome.
A reinforcement learning agent for a car sandboxed in a safe operating regime might merely learn to always accelerate at full speed, because its safety net will handle the rest to brake on time before impact.
Safe AI for CPS \cite{DBLP:conf/itc/FultonP18} can exploit the provably correct model monitoring and shielding technique ModelPlex \cite{DBLP:journals/fmsd/MitschP16} to identify when and how to interfere with an AI's decision in order to keep the CPS safe \cite{DBLP:conf/aaai/FultonP18}.
This approach even works for (limited) uncertainty about the physical model \cite{DBLP:conf/tacas/FultonP19}.
The resulting guarantees of Safe AI for CPS could also help overcome the explainability crisis that hinders the safe use of AI in railway applications \cite{DBLP:conf/isola/SeisenbergerBFF22}.

\section{Some Successful Intersymbolic AI Combinations} \label{sec:intersymbolicAI-survey}

While the idea of intersymbolic AI combining symbolic AI with subsymbolic AI is general, some successful combinations may help illustrate the generality and flexibility of intersymbolic AI.
Even if the respective intersymbolic AI features are remarkably different, the approaches are still discussed in groups to better tease out how similarities in goals may nevertheless manifest in remarkably different ways of forming intersymbolic AI out of different symbolic AI and subsymbolic AI combinations.
Intersymbolic AI is intentionally meant to be understood broadly and not limited solely to continuations of the following list of intersymbolic AI successes.

\paragraph{Safe AI in CPS.}
As explained in \rref{sec:IntersymbolicAIinCPS}, there is a clear and nontrivial need to make the use of AI safe in cyber-physical systems safe.
Safe AI in CPS \cite{DBLP:conf/itc/FultonP18,DBLP:conf/qest/Platzer19} has been made possible for the case of reinforcement learning with Justified Speculative Control \cite{DBLP:conf/aaai/FultonP18,DBLP:conf/tacas/FultonP19} as well as for neural network control systems with VerSAILLE \cite{DBLP:journals/corr/abs-2402-10998}.
In both cases, the symbolic AI technique of hybrid systems theorem proving for safe CPS \cite{Platzer10,Platzer18} has been combined with the subsymbolic AI of reinforcement learning for learning from active experimentation \cite{SuttonBarto18} or neural networks for learning from experimental data \cite{RusselNorvig:21}, respectively, which are interlinked with the symbolic AI technique of shielding via ModelPlex to overcome action discrepancies between known safe and possibly unsafe outcomes \cite{DBLP:journals/fmsd/MitschP16}.
Besides working for reinforcement learning in CPS and neural network control systems, respectively, the difference is that safety combines offline and online analysis elements \cite{DBLP:conf/aaai/FultonP18,DBLP:conf/tacas/FultonP19} or is entirely conducted offline before launch of the system \cite{DBLP:journals/corr/abs-2402-10998}. 
The neural network control systems verification technique also uses additional AI techniques of complete arithmetic linearization techniques \cite{DBLP:journals/corr/abs-2402-10998} to scale symbolic AI arithmetic verification to the scale of subsymbolic AI neural networks for control.
Intersymbolic AI in the form of combinations of the subsymbolic AI techniques of neural network path tracking and model-predictive control with symbolic AI hybrid systems theorem proving has also been used for safe waypoint following vehicles \cite{DBLP:journals/csyl/LinMPD22} using some of the Safe AI for CPS ideas.

\paragraph{Automatic Symbolic AI for CPS Proofs.}
Two very different intersymbolic AI approaches both support totally different ends of the spectrum of making symbolic AI for CPS proof search automatic \cite{Platzer10,Platzer10}.
One concerns Pegasus, the automatic invariant generator for differential equations \cite{DBLP:journals/fmsd/SogokonMTCP22}.
The other concerns CESAR, the automatic control envelope synthesizer for hybrid systems \cite{DBLP:conf/tacas/KabraLMP24}.
Pegasus automates the inner loop of the sophisticated mathematical task of generating invariants for differential equations, with which differential equations verification becomes decidable \cite{DBLP:journals/jacm/PlatzerT20}.
CESAR automates the creative control task of finding maximally flexible hybrid systems control constraints that make the hybrid system with its controllers and differential equations safe to begin with.

Even more different than the part of the safety verification task where they are used are the symbolic AI and subsymbolic AI techniques that are combined to make them happen.
Both Pegasus and CESAR share the use of symbolic AI ideas from logic for hybrid systems with differential equations \cite{Platzer10,Platzer18}.
But Pegasus also uses the symbolic concepts of first integrals of differential equations, which represent conserved quantities, and Darboux polynomials whose Lie derivatives are polynomial multiples of themselves.
CESAR, instead, uses the symbolic AI concepts of hybrid systems game theory to characterize optimal solutions \cite{DBLP:journals/tocl/Platzer15,Platzer18}.
Pegasus uses subsymbolic principles of eigensystems for linear differential equations with numerical sum-of-squares programming or linear programming to find parameter choices for nonlinear differential equation invariants,
while CESAR uses systematic underapproximation refinements and arithmetic simplification heuristics.

\paragraph{Learning to Learn and Teach Theorems.}

Proving theorems is hard and learning to prove theorems is not only difficult, but also made impossible by the acutely limited number of theorems and proofs to learn from, compared to the absurd data hunger of machine learning techniques.
That is why the Looprl prover \cite{DBLP:conf/nips/LaurentP22} circumvents the issue by providing both a student agent that learns how to prove theorems and a teacher agent that learns how to pose effective theorem proving challenges that the student agent can learn to prove theorems from.
Both use an AlphaZero agent \cite{DBLP:journals/nature/SilverSSAHGHBLB17} with Monte-Carlo Tree Search for a symbolic AI nondeterministic programming language to describe very high level generic proof search strategies (such as: ``find an invariant that proves the postcondition of a while loop program'') refined by subsymbolic AI deep neural networks trained via the AlphaZero agent.

\section{Conclusions and Outlook} \label{sec:Conclusion}

The most important observation in this perspective piece is the characterization of the different features of symbolic AI and subsymbolic AI and explanations where they come from, from the symbolic and subsymbolic nature respectively, as well as an analysis of their complementary strengths and weaknesses.
The most important contribution is the call to action in this new field that is here coined \emph{intersymbolic AI} combining symbolic AI and subsymbolic AI to increase the effectiveness of AI likened to the way that the combination of conscious and subconscious thought increases the effectiveness of human thought.
In vastly different applications, a wide variety of entirely different combinations of remarkably different symbolic AI and remarkably different subsymbolic AI techniques to form Intersymbolic AI highlight showcases for the power and potential of Intersymbolic AI.
It is this scientific diversity alongside the common underlying methodological metaprinciple captured in Intersymbolic AI that gives this perspective piece the biggest significance.
Future work abounds in the hope that an equally wide range of scientists notice the even wider potential of the use and advances of intersymbolic AI.

\paragraph*{Acknowledgment.}
We thank Jonathan Laurent, Noah Abou El Wafa, Alexander Weigl, Peter Sanders, Bernhard Beckert and Veit Hagenmeyer for their feedback.

\renewcommand{\doi}[1]{doi: \href{https://doi.org/#1}{\nolinkurl{#1}}}
\bibliographystyle{splncs04}
\bibliography{platzer,bibliography}

\begin{thebibliography}{100}
\providecommand{\url}[1]{\texttt{#1}}
\providecommand{\urlprefix}{URL }
\providecommand{\doi}[1]{https://doi.org/#1}

\bibitem{DBLP:journals/corr/AbadiABBCCCDDDG16}
Abadi, M., Agarwal, A., Barham, P., Brevdo, E., Chen, Z., Citro, C., Corrado,
  G.S., Davis, A., Dean, J., Devin, M., Ghemawat, S., Goodfellow, I.J., Harp,
  A., Irving, G., Isard, M., Jia, Y., J{\'{o}}zefowicz, R., Kaiser, L., Kudlur,
  M., Levenberg, J., Man{\'{e}}, D., Monga, R., Moore, S., Murray, D.G., Olah,
  C., Schuster, M., Shlens, J., Steiner, B., Sutskever, I., Talwar, K., Tucker,
  P.A., Vanhoucke, V., Vasudevan, V., Vi{\'{e}}gas, F.B., Vinyals, O., Warden,
  P., Wattenberg, M., Wicke, M., Yu, Y., Zheng, X.: {TensorFlow}: Large-scale
  machine learning on heterogeneous distributed systems. CoRR
  \textbf{abs/1603.04467} (2016)

\bibitem{Alur15}
Alur, R.: Principles of Cyber-Physical Systems. MIT Press, Cambridge (2015)

\bibitem{DBLP:journals/corr/AmodeiOSCSM16}
Amodei, D., Olah, C., Steinhardt, J., Christiano, P.F., Schulman, J.,
  Man{\'{e}}, D.: Concrete problems in {AI} safety. CoRR
  \textbf{abs/1606.06565} (2016)

\bibitem{DBLP:journals/iandc/Angluin87}
Angluin, D.: Learning regular sets from queries and counterexamples. Inf.
  Comput.  \textbf{75}(2),  87--106 (1987)

\bibitem{DBLP:journals/jacm/AtseriasM20}
Atserias, A., M{\"{u}}ller, M.: Automating resolution is {NP}-hard. J. {ACM}
  \textbf{67}(5),  31:1--31:17 (2020). \doi{10.1145/3409472}

\bibitem{DBLP:journals/bams/Avigad24}
Avigad, J.: Mathematics and the formal turn. Bull. Amer. Math. Soc.
  \textbf{61},  225--240 (2024). \doi{10.1090/bull/1832}

\bibitem{DBLP:journals/tocl/AvigadDGR07}
Avigad, J., Donnelly, K., Gray, D., Raff, P.: A formally verified proof of the
  prime number theorem. {ACM} Trans. Comput. Log.  \textbf{9}(1), ~2 (2007).
  \doi{10.1145/1297658.1297660}, \url{https://doi.org/10.1145/1297658.1297660}

\bibitem{DBLP:books/daglib/0041477}
Baader, F., Horrocks, I., Lutz, C., Sattler, U.: An Introduction to Description
  Logic. Cambridge University Press (2017)

\bibitem{DBLP:journals/logcom/BachmairG94}
Bachmair, L., Ganzinger, H.: Rewrite-based equational theorem proving with
  selection and simplification. J. Log. Comput.  \textbf{4}(3),  217--247
  (1994). \doi{10.1093/logcom/4.3.217}

\bibitem{DBLP:conf/icml/BansalLRSW19}
Bansal, K., Loos, S.M., Rabe, M.N., Szegedy, C., Wilcox, S.: {HOList}: An
  environment for machine learning of higher order logic theorem proving. In:
  Chaudhuri, K., Salakhutdinov, R. (eds.) Proceedings of the 36th International
  Conference on Machine Learning, {ICML} 2019, 9-15 June 2019, Long Beach,
  California, {USA}. Proceedings of Machine Learning Research, vol.~97, pp.
  454--463. {PMLR} (2019)

\bibitem{DBLP:series/faia/336}
Biere, A., Heule, M., van Maaren, H., Walsh, T. (eds.): Handbook of
  Satisfiability - Second Edition, Frontiers in Artificial Intelligence and
  Applications, vol.~336. {IOS} Press (2021). \doi{10.3233/FAIA336}

\bibitem{DBLP:conf/aaai/BoochFHKLLLMMRS21}
Booch, G., Fabiano, F., Horesh, L., Kate, K., Lenchner, J., Linck, N.,
  Loreggia, A., Murugesan, K., Mattei, N., Rossi, F., Srivastava, B.: Thinking
  fast and slow in {AI}. In: Thirty-Fifth {AAAI} Conference on Artificial
  Intelligence, {AAAI} 2021, Thirty-Third Conference on Innovative Applications
  of Artificial Intelligence, {IAAI} 2021, The Eleventh Symposium on
  Educational Advances in Artificial Intelligence, {EAAI} 2021, Virtual Event,
  February 2-9, 2021. pp. 15042--15046. {AAAI} Press (2021).
  \doi{10.1609/AAAI.V35I17.17765}

\bibitem{BrachmanLevesque14}
Brachman, R., Levesque, H.: Knowledge Representation and Reasoning. Morgan
  Kaufmann (2014)

\bibitem{Brewka91}
Brewka, G.: Nonmonotonic Reasoning. Cambridge Univ. Press (1991)

\bibitem{DBLP:journals/cacm/ClarkeES09}
Clarke, E.M., Emerson, E.A., Sifakis, J.: Model checking: algorithmic
  verification and debugging. Commun. ACM  \textbf{52}(11),  74--84 (2009).
  \doi{10.1145/1592761.1592781}

\bibitem{DBLP:conf/stoc/Cook71}
Cook, S.A.: The complexity of theorem-proving procedures. In: Harrison, M.A.,
  Banerji, R.B., Ullman, J.D. (eds.) STOC. pp. 151--158. ACM, New York (1971).
  \doi{10.1145/800157.805047}

\bibitem{DBLP:conf/ijcai/CropperDM20}
Cropper, A., Dumancic, S., Muggleton, S.H.: Turning 30: New ideas in inductive
  logic programming. In: Bessiere, C. (ed.) Proceedings of the Twenty-Ninth
  International Joint Conference on Artificial Intelligence, {IJCAI} 2020. pp.
  4833--4839. ijcai.org (2020). \doi{10.24963/IJCAI.2020/673},
  \url{https://doi.org/10.24963/ijcai.2020/673}

\bibitem{Eco76}
Eco, U.: A Theory of Semiotics. Advances in Semiotics, Indiana Univ. Press
  (1976)

\bibitem{DBLP:journals/jacm/EiterG95}
Eiter, T., Gottlob, G.: The complexity of logic-based abduction. J. {ACM}
  \textbf{42}(1),  3--42 (1995). \doi{10.1145/200836.200838}

\bibitem{DBLP:journals/cacm/Feigenbaum96}
Feigenbaum, E.A.: How the "what" becomes the "how" - turing award lecture.
  Commun. {ACM}  \textbf{39}(5),  97--104 (1996). \doi{10.1145/229459.229471}

\bibitem{DBLP:conf/ijcai/FikesN71}
Fikes, R., Nilsson, N.J.: {STRIPS:} {A} new approach to the application of
  theorem proving to problem solving. In: Cooper, D.C. (ed.) Proceedings of the
  2nd International Joint Conference on Artificial Intelligence. London, UK,
  September 1-3, 1971. pp. 608--620. William Kaufmann (1971)

\bibitem{Fitting96a}
Fitting, M.: First-Order Logic and Automated Theorem Proving. Springer, New
  York, 2nd edn. (1996)

\bibitem{Frege79}
Frege, G.: Begriffsschrift, eine der arithmetischen nachgebildete
  {Formelsprache} des reinen {Denkens}. Verlag von Louis Nebert, Halle (1879).
  \doi{10.1007/978-3-662-45011-6}

\bibitem{DBLP:conf/itc/FultonP18}
Fulton, N., Platzer, A.: Safe {AI} for {CPS}. In: {IEEE} International Test
  Conference, {ITC} 2018, Phoenix, AZ, USA, October 29 - Nov. 1, 2018.
  pp.~1--7. IEEE (2018). \doi{10.1109/TEST.2018.8624774}

\bibitem{DBLP:conf/aaai/FultonP18}
Fulton, N., Platzer, A.: Safe reinforcement learning via formal methods: Toward
  safe control through proof and learning. In: McIlraith, S.A., Weinberger,
  K.Q. (eds.) AAAI. pp. 6485--6492. {AAAI} Press (2018)

\bibitem{DBLP:conf/tacas/FultonP19}
Fulton, N., Platzer, A.: Verifiably safe off-model reinforcement learning. In:
  Vojnar, T., Zhang, L. (eds.) TACAS, Part {I}. LNCS, vol. 11427, pp. 413--430.
  Springer (2019). \doi{10.1007/978-3-030-17462-0_28}

\bibitem{DBLP:conf/fat/GanguliHLABCCDD22}
Ganguli, D., Hernandez, D., Lovitt, L., Askell, A., Bai, Y., Chen, A., Conerly,
  T., DasSarma, N., Drain, D., Elhage, N., Showk, S.E., Fort, S.,
  Hatfield{-}Dodds, Z., Henighan, T., Johnston, S., Jones, A., Joseph, N.,
  Kernian, J., Kravec, S., Mann, B., Nanda, N., Ndousse, K., Olsson, C.,
  Amodei, D., Brown, T.B., Kaplan, J., McCandlish, S., Olah, C., Amodei, D.,
  Clark, J.: Predictability and surprise in large generative models. In: FAccT
  '22: 2022 {ACM} Conference on Fairness, Accountability, and Transparency,
  Seoul, Republic of Korea, June 21 - 24, 2022. pp. 1747--1764. {ACM} (2022).
  \doi{10.1145/3531146.3533229}

\bibitem{DBLP:series/cogtech/GarcezLG2009}
d'Avila Garcez, A.S., Lamb, L.C., Gabbay, D.M.: Neural-Symbolic Cognitive
  Reasoning. Cognitive Technologies, Springer (2009).
  \doi{10.1007/978-3-540-73246-4}

\bibitem{Geffner:thesis}
Geffner, H.: Default reasoning: causal and conditional theories. MIT Press
  (1992)

\bibitem{DBLP:series/synthesis/2013Geffner}
Geffner, H., Bonet, B.: A Concise Introduction to Models and Methods for
  Automated Planning. Synthesis Lectures on Artificial Intelligence and Machine
  Learning, Morgan {\&} Claypool Publishers (2013).
  \doi{10.2200/S00513ED1V01Y201306AIM022}

\bibitem{Gentzen35I}
Gentzen, G.: Untersuchungen {\"u}ber das logische {Schlie{\ss}en} {I}. Math.
  Zeit.  \textbf{39}(2),  176--210 (1935). \doi{10.1007/BF01201353}

\bibitem{GhallabNT16}
Ghallab, M., Nau, D., Traverso, P.: Automated Planning and Acting. Cambridge
  Univ. Press (2016)

\bibitem{Goedel30}
G{\"o}del, K.: Die {Vollst{\"a}ndigkeit} der {Axiome} des logischen
  {Funktionenkalk{\"u}ls}. Monatshefte Math. Phys.  \textbf{37},  349--360
  (1930). \doi{10.1007/BF01696781}

\bibitem{Godel32}
G{\"o}del, K.: Zum intuitionistischen {Aussagenkalk{\"u}l}. Anzeiger Akademie
  der Wissenschaften Wien  \textbf{69},  65--66 (1932)

\bibitem{Gonthier05}
Gonthier, G.: A computer-checked proof of the four colour theorem. Tech. Rep.
  hal-04034866, INRIA (2005), hAL report 2023

\bibitem{DBLP:conf/itp/GonthierAABCGRMOBPRSTT13}
Gonthier, G., Asperti, A., Avigad, J., Bertot, Y., Cohen, C., Garillot, F.,
  Roux, S.L., Mahboubi, A., O'Connor, R., Biha, S.O., Pasca, I., Rideau, L.,
  Solovyev, A., Tassi, E., Th{\'{e}}ry, L.: A machine-checked proof of the odd
  order theorem. In: Blazy, S., Paulin{-}Mohring, C., Pichardie, D. (eds.)
  Interactive Theorem Proving - 4th International Conference, {ITP} 2013,
  Rennes, France, July 22-26, 2013. Proceedings. LNCS, vol.~7998, pp. 163--179.
  Springer (2013). \doi{10.1007/978-3-642-39634-2_14}

\bibitem{DBLP:journals/cacm/GoodfellowPMXWO20}
Goodfellow, I.J., Pouget{-}Abadie, J., Mirza, M., Xu, B., Warde{-}Farley, D.,
  Ozair, S., Courville, A.C., Bengio, Y.: Generative adversarial networks.
  Commun. {ACM}  \textbf{63}(11),  139--144 (2020). \doi{10.1145/3422622}

\bibitem{DBLP:conf/cav/GouwRBBH15}
de~Gouw, S., Rot, J., de~Boer, F.S., Bubel, R., H{\"{a}}hnle, R.: {OpenJDK's}
  {Java.utils.Collection.sort()} is broken: The good, the bad and the worst
  case. In: Kroening, D., Pasareanu, C.S. (eds.) Computer Aided Verification -
  27th International Conference, {CAV} 2015, San Francisco, CA, USA, July
  18-24, 2015, Proceedings, Part {I}. LNCS, vol.~9206, pp. 273--289. Springer
  (2015). \doi{10.1007/978-3-319-21690-4_16}

\bibitem{Hales17}
Hales, T.C., Adams, M., Bauer, G., Dang, D.T., Harrison, J., Hoang, T.L.,
  Kaliszyk, C., Magron, V., McLaughlin, S., Nguyen, T.T., Nguyen, T.Q., Nipkow,
  T., Obua, S., Pleso, J., Rute, J.M., Solovyev, A., Ta, A.H.T., Tran, T.N.,
  Trieu, D.T., Urban, J., Vu, K.K., Zumkeller, R.: A formal proof of the
  {Kepler} conjecture. Forum of Mathematics, Pi  \textbf{5}, ~e2 (2017).
  \doi{DOI: 10.1017/fmp.2017.1}

\bibitem{DBLP:books/sp/HS2018}
Hamadi, Y., Sais, L. (eds.): Handbook of Parallel Constraint Reasoning.
  Springer (2018). \doi{10.1007/978-3-319-63516-3}

\bibitem{Haugeland89}
Haugeland, J.: Artificial Intelligence: the very idea. {MIT} Press, USA (1989)

\bibitem{Hsu02}
Hsu, F.h.: Behind {Deep} {Blue}: building the computer that defeated the world
  chess champion. Princeton University Press, Princeton (2002)

\bibitem{DBLP:journals/sttt/JeanninGKSGMP17}
Jeannin, J., Ghorbal, K., Kouskoulas, Y., Schmidt, A., Gardner, R., Mitsch, S.,
  Platzer, A.: A formally verified hybrid system for safe advisories in the
  next-generation airborne collision avoidance system. STTT  \textbf{19}(6),
  717--741 (2017). \doi{10.1007/s10009-016-0434-1}

\bibitem{DBLP:conf/tacas/KabraLMP24}
Kabra, A., Laurent, J., Mitsch, S., Platzer, A.: {CESAR}: Control envelope
  synthesis via angelic refinements. In: Finkbeiner, B., Kov{\'{a}}cs, L.
  (eds.) TACAS. LNCS, vol. 14570, pp. 144--164. Springer (2024).
  \doi{10.1007/978-3-031-57246-3_9}

\bibitem{DBLP:journals/corr/abs-2001-08361}
Kaplan, J., McCandlish, S., Henighan, T., Brown, T.B., Chess, B., Child, R.,
  Gray, S., Radford, A., Wu, J., Amodei, D.: Scaling laws for neural language
  models. CoRR  \textbf{abs/2001.08361} (2020)

\bibitem{DBLP:journals/aim/Kautz22}
Kautz, H.A.: The third {AI} summer: {AAAI} {Robert S. Engelmore} memorial
  lecture. {AI} Mag.  \textbf{43}(1),  93--104 (2022).
  \doi{10.1609/AIMAG.V43I1.19122}

\bibitem{DBLP:journals/arcras/KwiatkowskaN022}
Kwiatkowska, M., Norman, G., Parker, D.: Probabilistic model checking and
  autonomy. Annu. Rev. Control. Robotics Auton. Syst.  \textbf{5},  385--410
  (2022). \doi{10.1146/annurev-control-042820-010947}

\bibitem{DBLP:conf/fedcsis/KwiatkowskaZ23}
Kwiatkowska, M., Zhang, X.: When to trust {AI:} advances and challenges for
  certification of neural networks. In: Ganzha, M., Maciaszek, L.A., Paprzycki,
  M., Slezak, D. (eds.) Proceedings of the 18th Conference on Computer Science
  and Intelligence Systems, FedCSIS 2023, Warsaw, Poland, September 17-20,
  2023. Annals of Computer Science and Information Systems, vol.~35, pp. 25--37
  (2023). \doi{10.15439/2023F2324}

\bibitem{DBLP:conf/nips/LaurentP22}
Laurent, J., Platzer, A.: Learning to find proofs and theorems by learning to
  refine search strategies. In: Koyejo, S., Mohamed, S., Agarwal, A., Belgrave,
  D., Cho, K., Oh, A. (eds.) Advances in Neural Information Processing Systems.
  vol.~35, p. 4843–4856. Curran Associates, Inc. (2022)

\bibitem{LeCunBD89}
LeCun, Y., Boser, B., Denker, J.S., Henderson, D., Howard, R.E., Hubbard, W.,
  Jackel, L.D.: Backpropagation applied to handwritten zip code recognition.
  Neural Computation  \textbf{1}(4),  541--551 (12 1989).
  \doi{10.1162/neco.1989.1.4.541}

\bibitem{LeeSeshia13}
Lee, E.A., Seshia, S.A.: Introduction to Embedded Systems --- A Cyber-Physical
  Systems Approach. Lulu.com (2013)

\bibitem{DBLP:journals/csyl/LinMPD22}
Lin, Q., Mitsch, S., Platzer, A., Dolan, J.M.: Safe and resilient practical
  waypoint-following for autonomous vehicles. {IEEE} Control. Syst. Lett.
  \textbf{6},  1574--1579 (2022). \doi{10.1109/LCSYS.2021.3125717}

\bibitem{Lukaszewicz90}
{\L{}}ukaszewicz, W.: Non-monotonic Reasoning. Ellis Horwood (1990)

\bibitem{GaelFR24}
Martin, G.M., Frazier, D.T., Robert, C.P.: Computing {Bayes}: From then `til
  now. Statistical Science  \textbf{39}(1),  3 -- 19 (2024).
  \doi{10.1214/22-STS876}

\bibitem{MartinLof80}
Martin-L{\"o}f, P.: Constructive mathematics and computer programming. In:
  Logic, Methodology and Philosophy of Science VI. pp. 153--175. North-Holland
  (1980)

\bibitem{Marwedel21}
Marwedel, P.: Embedded System Design: Embedded Systems Foundations of
  Cyber-Physical Systems, and the Internet of Things. Springer, 4 edn. (2021).
  \doi{10.1007/978-3-030-60910-8}

\bibitem{DBLP:journals/jar/McCune97}
McCune, W.: Solution of the {Robbins} problem. J. Autom. Reason.
  \textbf{19}(3),  263--276 (1997). \doi{10.1023/A:1005843212881}

\bibitem{Mitchell:240}
Mitchell, T.: Machine Learning. McGraw-Hill (1997)

\bibitem{Mitra21}
Mitra, S.: Verifying Cyber-Physical Systems: A Path to Safe Autonomy. MIT Press
  (2021)

\bibitem{DBLP:journals/fmsd/MitschP16}
Mitsch, S., Platzer, A.: {ModelPlex}: Verified runtime validation of verified
  cyber-physical system models. Form. Methods Syst. Des.  \textbf{49}(1-2),
  33--74 (2016). \doi{10.1007/s10703-016-0241-z}, special issue of selected
  papers from RV'14

\bibitem{DBLP:journals/csur/MitschPRS15}
Mitsch, S., Platzer, A., Retschitzegger, W., Schwinger, W.: Logic-based
  modeling approaches for qualitative and hybrid reasoning in dynamic spatial
  systems. {ACM} Comput. Surv.  \textbf{48}(1),  3:1--3:40 (2015).
  \doi{10.1145/2764901}

\bibitem{vonNeumannMorgenstern}
von Neumann, J., Morgenstern, O.: Theory of Games and Economic Behavior.
  Princeton Univ. Press, 3rd edn. (1955)

\bibitem{DBLP:journals/ibmrd/NewellSS58}
Newell, A., Shaw, J.C., Simon, H.A.: Chess-playing programs and the problem of
  complexity. {IBM} J. Res. Dev.  \textbf{2}(4),  320--335 (1958).
  \doi{10.1147/RD.24.0320}

\bibitem{DBLP:conf/ifip/NewellSS59}
Newell, A., Shaw, J.C., Simon, H.A.: Report on a general problem-solving
  program. In: Information Processing, Proceedings of the 1st International
  Conference on Information Processing, UNESCO, Paris 15-20 June 1959. pp.
  256--264. {UNESCO} (Paris) (1959)

\bibitem{DBLP:journals/tit/NewellS56}
Newell, A., Simon, H.A.: The logic theory machine-a complex information
  processing system. {IRE} Trans. Inf. Theory  \textbf{2}(3),  61--79 (1956).
  \doi{10.1109/TIT.1956.1056797}

\bibitem{DBLP:journals/cacm/NewellS76}
Newell, A., Simon, H.A.: Computer science as empirical inquiry: Symbols and
  search. Commun. {ACM}  \textbf{19}(3),  113--126 (1976).
  \doi{10.1145/360018.360022}

\bibitem{DBLP:journals/jacm/NieuwenhuisOT06}
Nieuwenhuis, R., Oliveras, A., Tinelli, C.: Solving {SAT} and {SAT} modulo
  theories: From an abstract {Davis}--{Putnam}--{Logemann}--{Loveland}
  procedure to {DPLL}({\emph{t}}). J. {ACM}  \textbf{53}(6),  937--977 (2006).
  \doi{10.1145/1217856.1217859}

\bibitem{DBLP:conf/sp/NoppelPW23}
Noppel, M., Peter, L., Wressnegger, C.: Disguising attacks with
  explanation-aware backdoors. In: 44th {IEEE} Symposium on Security and
  Privacy, {SP} 2023, San Francisco, CA, USA, May 21-25, 2023. pp. 664--681.
  {IEEE} (2023). \doi{10.1109/SP46215.2023.10179308}

\bibitem{DBLP:conf/aaai/PaliwalLRBS20}
Paliwal, A., Loos, S.M., Rabe, M.N., Bansal, K., Szegedy, C.: Graph
  representations for higher-order logic and theorem proving. In: The
  Thirty-Fourth {AAAI} Conference on Artificial Intelligence, {AAAI} 2020, The
  Thirty-Second Innovative Applications of Artificial Intelligence Conference,
  {IAAI} 2020, The Tenth {AAAI} Symposium on Educational Advances in Artificial
  Intelligence, {EAAI} 2020, New York, NY, USA, February 7-12, 2020. pp.
  2967--2974. {AAAI} Press (2020). \doi{10.1609/AAAI.V34I03.5689}

\bibitem{DBLP:conf/nips/PaszkeGMLBCKLGA19}
Paszke, A., Gross, S., Massa, F., Lerer, A., Bradbury, J., Chanan, G., Killeen,
  T., Lin, Z., Gimelshein, N., Antiga, L., Desmaison, A., K{\"{o}}pf, A., Yang,
  E.Z., DeVito, Z., Raison, M., Tejani, A., Chilamkurthy, S., Steiner, B.,
  Fang, L., Bai, J., Chintala, S.: {PyTorch}: An imperative style,
  high-performance deep learning library. In: Wallach, H.M., Larochelle, H.,
  Beygelzimer, A., d'Alch{\'{e}}{-}Buc, F., Fox, E.B., Garnett, R. (eds.)
  Advances in Neural Information Processing Systems 32: Annual Conference on
  Neural Information Processing Systems 2019, NeurIPS 2019, December 8-14,
  2019, Vancouver, BC, Canada. pp. 8024--8035 (2019)

\bibitem{DBLP:books/daglib/0068933}
Pearl, J.: Heuristics - intelligent search strategies for computer problem
  solving. Addison-Wesley series in artificial intelligence, Addison-Wesley
  (1984)

\bibitem{DBLP:books/daglib/0066829}
Pearl, J.: Probabilistic reasoning in intelligent systems - networks of
  plausible inference. Morgan Kaufmann series in representation and reasoning,
  Morgan Kaufmann (1989)

\bibitem{DBLP:journals/cogsci/PetersKB22}
Peters, U., Krauss, A., Braganza, O.: Generalization bias in science. Cogn.
  Sci.  \textbf{46}(9) (2022). \doi{10.1111/COGS.13188}

\bibitem{Platzer10}
Platzer, A.: Logical Analysis of Hybrid Systems: Proving Theorems for Complex
  Dynamics. Springer, Heidelberg (2010). \doi{10.1007/978-3-642-14509-4}

\bibitem{DBLP:conf/lics/Platzer12a}
Platzer, A.: Logics of dynamical systems. In: LICS. pp. 13--24. IEEE, Los
  Alamitos (2012). \doi{10.1109/LICS.2012.13}

\bibitem{DBLP:journals/tocl/Platzer15}
Platzer, A.: Differential game logic. {ACM} Trans. Comput. Log.
  \textbf{17}(1),  1:1--1:51 (2015). \doi{10.1145/2817824}

\bibitem{Platzer18}
Platzer, A.: Logical Foundations of Cyber-Physical Systems. Springer, Cham
  (2018). \doi{10.1007/978-3-319-63588-0}

\bibitem{DBLP:conf/qest/Platzer19}
Platzer, A.: The logical path to autonomous cyber-physical systems. In: Parker,
  D., Wolf, V. (eds.) QEST. LNCS, vol. 11785, pp. 25--33. Springer (2019).
  \doi{10.1007/978-3-030-30281-8_2}

\bibitem{DBLP:conf/hybrid/PlatzerC07}
Platzer, A., Clarke, E.M.: The image computation problem in hybrid systems
  model checking. In: Bemporad, A., Bicchi, A., Buttazzo, G. (eds.) HSCC. LNCS,
  vol.~4416, pp. 473--486. Springer, Berlin (2007).
  \doi{10.1007/978-3-540-71493-4_37}

\bibitem{DBLP:journals/jacm/PlatzerT20}
Platzer, A., Tan, Y.K.: Differential equation invariance axiomatization. J. ACM
   \textbf{67}(1),  6:1--6:66 (2020). \doi{10.1145/3380825}

\bibitem{Popper62}
Popper, K.R.: Conjectures and Refutations: The Growth of Scientific Knowledge.
  Routledge, London, England (1962)

\bibitem{DBLP:journals/jsyml/Quine50}
Quine, W.V.: On natural deduction. J. Symb. Log.  \textbf{15}(2),  93--102
  (1950)

\bibitem{DBLP:conf/cade/RegerTV15}
Reger, G., Tishkovsky, D., Voronkov, A.: Cooperating proof attempts. In: Felty,
  A.P., Middeldorp, A. (eds.) Automated Deduction - {CADE-25} - 25th
  International Conference on Automated Deduction, Berlin, Germany, August 1-7,
  2015, Proceedings. LNCS, vol.~9195, pp. 339--355. Springer (2015).
  \doi{10.1007/978-3-319-21401-6_23}

\bibitem{DBLP:conf/birthday/Reiter91}
Reiter, R.: The frame problem in the situation calculus: {A} simple solution
  (sometimes) and a completeness result for goal regression. In: Lifschitz, V.
  (ed.) Artificial and Mathematical Theory of Computation, Papers in Honor of
  John McCarthy on the occasion of his sixty-fourth birthday. pp. 359--380.
  Academic Press / Elsevier (1991). \doi{10.1016/B978-0-12-450010-5.50026-8}

\bibitem{DBLP:conf/kr/Reiter96}
Reiter, R.: Natural actions, concurrency and continuous time in the situation
  calculus. In: Aiello, L.C., Doyle, J., Shapiro, S.C. (eds.) Proceedings of
  the Fifth International Conference on Principles of Knowledge Representation
  and Reasoning (KR'96), Cambridge, Massachusetts, USA, November 5-8, 1996. pp.
  2--13. Morgan Kaufmann (1996)

\bibitem{DBLP:conf/icfem/RenshawLP11}
Renshaw, D.W., Loos, S.M., Platzer, A.: Distributed theorem proving for
  distributed hybrid systems. In: Qin, S., Qiu, Z. (eds.) ICFEM. LNCS,
  vol.~6991, pp. 356--371. Springer (2011). \doi{10.1007/978-3-642-24559-6_25}

\bibitem{DBLP:conf/stoc/RezendeGNPR021}
de~Rezende, S.F., G{\"{o}}{\"{o}}s, M., Nordstr{\"{o}}m, J., Pitassi, T.,
  Robere, R., Sokolov, D.: Automating algebraic proof systems is {NP}-hard. In:
  Khuller, S., Williams, V.V. (eds.) {STOC} '21: 53rd Annual {ACM} {SIGACT}
  Symposium on Theory of Computing, Virtual Event, Italy, June 21-25, 2021. pp.
  209--222. {ACM} (2021). \doi{10.1145/3406325.3451080}

\bibitem{DBLP:books/el/RobinsonV01}
Robinson, J.A., Voronkov, A. (eds.): Handbook of Automated Reasoning. MIT Press
  (2001)

\bibitem{RusselNorvig:21}
Russel, S., Norvig, P.: Artificial Intelligence: a Modern Approach. Pearson, 4
  edn. (2021)

\bibitem{SchaefferBBKMLLS07}
Schaeffer, J., Burch, N., Bj{\"o}rnsson, Y., Kishimoto, A., M{\"u}ller, M.,
  Lake, R., Lu, P., Sutphen, S.: Checkers is solved. Science
  \textbf{317}(5844),  1518--1522 (2007). \doi{10.1126/science.1144079}

\bibitem{DBLP:conf/sat/SchreiberS21}
Schreiber, D., Sanders, P.: Scalable {SAT} solving in the cloud. In: Li, C.,
  Many{\`{a}}, F. (eds.) Theory and Applications of Satisfiability Testing -
  {SAT} 2021 - 24th International Conference, Barcelona, Spain, July 5-9, 2021,
  Proceedings. LNCS, vol. 12831, pp. 518--534. Springer (2021).
  \doi{10.1007/978-3-030-80223-3_35}

\bibitem{DBLP:conf/isola/SeisenbergerBFF22}
Seisenberger, M., ter Beek, M.H., Fan, X., Ferrari, A., Haxthausen, A.E.,
  James, P., Lawrence, A., Luttik, B., van~de Pol, J., Wimmer, S.: Safe and
  secure future {AI}-driven railway technologies: Challenges for formal methods
  in railway. In: Margaria, T., Steffen, B. (eds.) Leveraging Applications of
  Formal Methods, Verification and Validation. Practice, ISoLA. LNCS, vol.
  13704, pp. 246--268. Springer (2022). \doi{10.1007/978-3-031-19762-8_20}

\bibitem{DBLP:books/daglib/0095085}
Shanahan, M.: Solving the frame problem - a mathematical investigation of the
  common sense law of inertia. {MIT} Press (1997)

\bibitem{DBLP:journals/corr/abs-1712-01815}
Silver, D., Hubert, T., Schrittwieser, J., Antonoglou, I., Lai, M., Guez, A.,
  Lanctot, M., Sifre, L., Kumaran, D., Graepel, T., Lillicrap, T.P., Simonyan,
  K., Hassabis, D.: Mastering {Chess} and {Shogi} by self-play with a general
  reinforcement learning algorithm. CoRR  \textbf{abs/1712.01815} (2017)

\bibitem{DBLP:journals/nature/SilverSSAHGHBLB17}
Silver, D., Schrittwieser, J., Simonyan, K., Antonoglou, I., Huang, A., Guez,
  A., Hubert, T., Baker, L., Lai, M., Bolton, A., Chen, Y., Lillicrap, T.P.,
  Hui, F., Sifre, L., van~den Driessche, G., Graepel, T., Hassabis, D.:
  Mastering the game of {Go} without human knowledge. Nat.  \textbf{550}(7676),
   354--359 (2017). \doi{10.1038/NATURE24270}

\bibitem{DBLP:journals/fmsd/SogokonMTCP22}
Sogokon, A., Mitsch, S., Tan, Y.K., Cordwell, K., Platzer, A.: Pegasus: Sound
  continuous invariant generation. Form. Methods Syst. Des.  \textbf{58}(1),
  5--41 (2022). \doi{10.1007/s10703-020-00355-z}, special issue for selected
  papers from FM'19

\bibitem{Sowa00}
Sowa, J.F.: Knowledge Representation: logical, philosophical, and computational
  foundations. Brooks/Cole, Pacific Grove, CA (2000)

\bibitem{DBLP:conf/bica/SukhobokovGC19}
Sukhobokov, A.A., Gapanyuk, Y.E., Chernenkiy, V.M.: Consciousness and
  subconsciousness as a means of {AGI's} and narrow {AI's} integration. In:
  Samsonovich, A.V. (ed.) Biologically Inspired Cognitive Architectures 2019.
  Advances in Intelligent Systems and Computing, vol.~948, pp. 515--520.
  Springer (2019). \doi{10.1007/978-3-030-25719-4_66}

\bibitem{SuttonBarto18}
Sutton, R.S., Barto, A.G.: Reinforcement Learning: An Introduction. Bradford
  Books, 2 edn. (2018)

\bibitem{DBLP:journals/corr/abs-2402-10998}
Teuber, S., Mitsch, S., Platzer, A.: Provably safe neural network controllers
  via differential dynamic logic. CoRR  \textbf{abs/2402.10998} (2024)

\bibitem{DBLP:journals/tplp/Thielscher05}
Thielscher, M.: {FLUX:} {A} logic programming method for reasoning agents.
  Theory Pract. Log. Program.  \textbf{5}(4-5),  533--565 (2005).
  \doi{10.1017/S1471068405002358}

\bibitem{TowellS94}
Towell, G.G., Shavlik, J.W.: Knowledge-based artificial neural networks.
  Artificial Intelligence  \textbf{70}(1),  119--165 (1994).
  \doi{10.1016/0004-3702(94)90105-8}

\bibitem{DBLP:journals/bsl/Urquhart95}
Urquhart, A.: The complexity of propositional proofs. Bull. Symb. Log.
  \textbf{1}(4),  425--467 (1995). \doi{10.2307/421131},
  \url{https://doi.org/10.2307/421131}

\bibitem{DBLP:conf/nips/VaswaniSPUJGKP17}
Vaswani, A., Shazeer, N., Parmar, N., Uszkoreit, J., Jones, L., Gomez, A.N.,
  Kaiser, L., Polosukhin, I.: Attention is all you need. In: Guyon, I., von
  Luxburg, U., Bengio, S., Wallach, H.M., Fergus, R., Vishwanathan, S.V.N.,
  Garnett, R. (eds.) Advances in Neural Information Processing Systems 30. pp.
  5998--6008 (2017)

\bibitem{DBLP:conf/itp/Wenzel13}
Wenzel, M.: Shared-memory multiprocessing for interactive theorem proving. In:
  Blazy, S., Paulin{-}Mohring, C., Pichardie, D. (eds.) Interactive Theorem
  Proving - 4th International Conference, {ITP} 2013, Rennes, France, July
  22-26, 2013. Proceedings. LNCS, vol.~7998, pp. 418--434. Springer (2013).
  \doi{10.1007/978-3-642-39634-2_30}

\end{thebibliography}
\end{document}